\documentclass[runningheads]{llncs}
\usepackage[T1]{fontenc}
\usepackage{graphicx}
\usepackage{hyperref}

\usepackage[misc,geometry]{ifsym}
\usepackage{booktabs}
\usepackage[most]{tcolorbox}

\definecolor{Emerald}{rgb}{0.31,0.78,0.47}

\usepackage{bbding}
\usepackage{multirow}
\usepackage{amsmath}
\usepackage{amssymb}
\usepackage{tabularx}
\usepackage[table]{xcolor}
\usepackage[font=small, labelfont=bf]{caption} 
\usepackage{subcaption}
\usepackage{orcidlink}

\begin{document}
\title{Evaluating Scoring Bias in LLM-as-a-Judge}

\author{
    Qingquan Li$^{\star \dagger}$\orcidlink{0009-0005-7605-1247},
    Shaoyu Dou$^\star$\orcidlink{0000-0002-8899-4788},
    Kailai Shao,
    Chao Chen,
    Haixiang Hu \Letter
}
\authorrunning{Q. Li et al.}
\institute{
Ant Group, Hangzhou, China \\
\email{qingquanl97@outlook.com, doushaoyu.dsy@antgroup.com, kailai.skl@antgroup.com,
    chixi.cc@antgroup.com, zengxian@antgroup.com}
}


\maketitle              

\begingroup
\renewcommand{\thefootnote}{}
\setcounter{footnote}{0}
\footnotetext{$^\star$Equal contribution. $^\dagger$Work done while at Ant Group}
\footnotetext{\textsuperscript{1}Our dataset is publicly available at \url{https://github.com/KMdsy/scoring_bias/}}
\endgroup

\begin{abstract}
The ``LLM-as-a-Judge'' paradigm, using Large Language Models (LLMs) as automated evaluators, is pivotal to LLM development, offering scalable feedback for complex tasks. However, the reliability of these judges is compromised by various biases. Existing research has heavily concentrated on biases in comparative evaluations. In contrast, scoring-based evaluations—which assign an absolute score and are often more practical in industrial applications—remain under-investigated.
To address this gap, we undertake the first dedicated examination of scoring bias in LLM judges. We shift the focus from biases tied to the evaluation targets to those originating from the scoring prompt itself. We formally define scoring bias and identify three novel, previously unstudied types: rubric order bias, score ID bias, and reference answer score bias.
We propose a comprehensive framework to quantify these biases, featuring a suite of multi-faceted metrics and an automatic data synthesis pipeline to create a tailored evaluation corpus\textsuperscript{1}. Our experiments empirically demonstrate that even the most advanced LLMs suffer from these substantial scoring biases. Our analysis yields actionable insights for designing more robust scoring prompts and mitigating these newly identified biases.
\keywords{LLM-as-a-Judge \and Bias.}
\end{abstract}

\section{Introduction}
The ``LLM-as-a-Judge''~\cite{gu2024survey,li2024generation} paradigm, where Large Language Models (LLMs) serve as automated evaluators, is pivotal to LLM development. These judges provide fine-grained assessments for complex tasks like factuality evaluation~\cite{rahman2025hallucination} and instruction-following~\cite{zeng2024evaluating}, and act as scalable proxies for human feedback in preference learning~\cite{bai2022constitutional,lee2024rlaif}. The increasing prominence of this paradigm makes its robustness a critical concern. Ideally, LLM judges should provide objective judgments, uninfluenced by irrelevant factors. However, recent work~\cite{park2024offsetbias} confirms that LLM judges exhibit various biases, undermining their consistency. Existing research has identified several such biases, including position bias~\cite{wang2024large}, self-preference bias~\cite{wataoka2024self}, length bias~\cite{wei2024systematic}, and so on.

\begin{figure}[!t]
    \centering
    \includegraphics[width=0.85\linewidth]{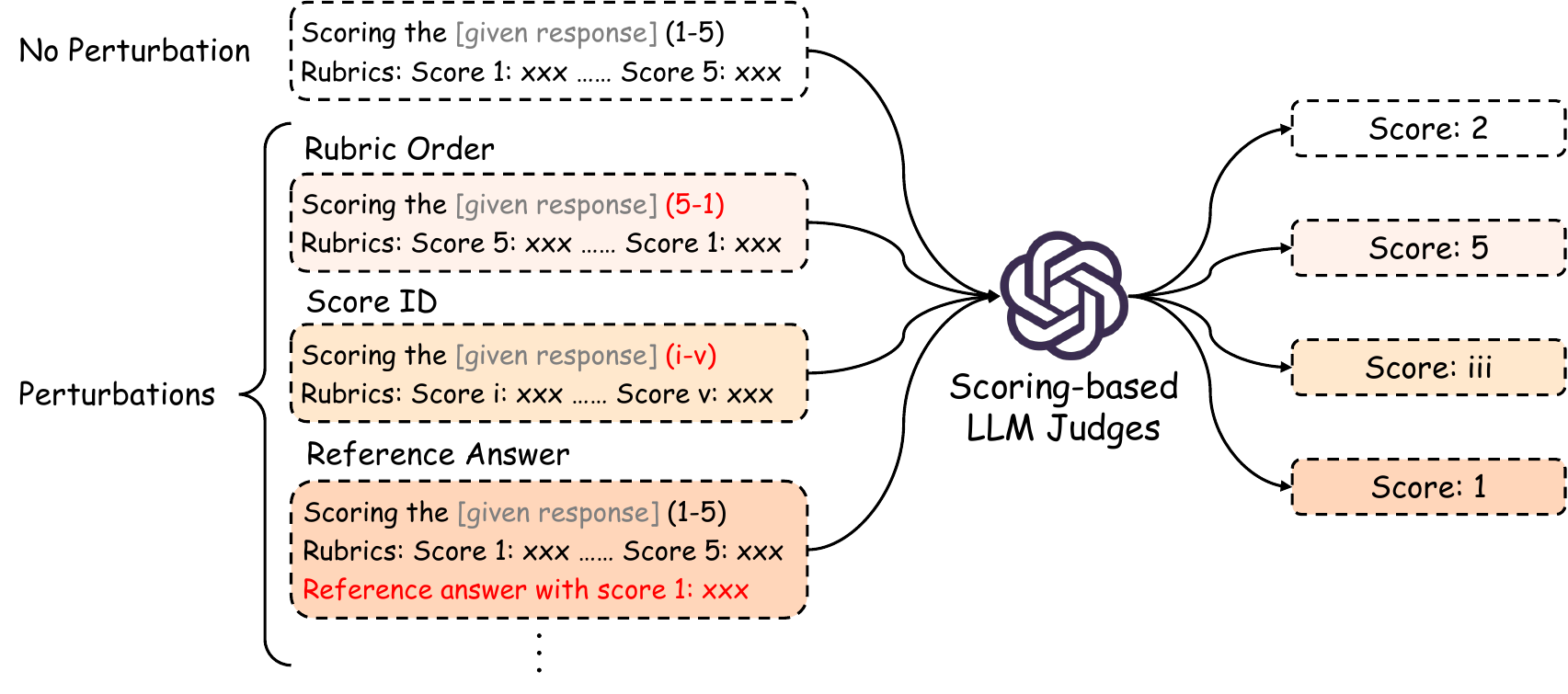}
    \caption{Even slight perturbations to LLM judges lead to inconsistent scores for the same response.}
    \label{fig:start_example}
\end{figure}

While existing studies have investigated these biases, pressing issues remain. Research has heavily concentrated on \textit{comparative} LLM judges, which determine the superior of two responses. However, \textit{scoring-based} evaluation, providing an absolute quality score for a single response, is often more practical and widely adopted. It is more intuitive and better suited for standardized, quantitative analysis, especially when a direct comparison object is unavailable~\cite{tripathi2025pairwisepointwiseevaluatingfeedback}. Consequently, an in-depth investigation into the biases of scoring-based judges is crucial. Research in this area remains unsystematic; while some issues like self-enhancement bias, refinement-aware bias~\cite{ye2024justice}, and judge model choice bias~\cite{wang2024large} have been identified, the exploration is far from comprehensive and leaves many potential biases unaddressed.

This research gap is not merely theoretical. Our work is motivated by a practical observation: when using a score-based LLM judge to evaluate a financial advisory assistant, we found that adding a single in-context example for a specific score level (e.g., an exemplar for a 4-point response) caused the absolute scores of other, unrelated samples to shift. This instability highlights a critical, unaddressed sensitivity to prompt examples and served as the primary catalyst for our investigation into other potential biases affecting score-based judges.

In this work, we consider the \textbf{scoring bias} as the measurable scoring shifts exhibited by an LLM-as-a-Judge when subjected to perturbations to its scoring prompt (as illustrated in Figure~\ref{fig:start_example}). Diverging from prior work, which has focused heavily on biases tied to the evaluation targets, our research investigates biases originating from other components of the scoring prompt. Specifically, we identify and define three novel and previously unstudied types of scoring bias: rubric order bias, score ID bias, and reference answer score bias. 

To identify and quantify these three biases, we propose a comprehensive evaluation framework. This framework is designed to be extensible, enabling the identification of various scoring biases beyond those explored in this paper. We conduct experiments on four existing LLM-as-a-Judge benchmarks and utilize a suite of metrics, categorized as stability metrics, accuracy metrics, and scoring tendency, to thoroughly quantify the impact of these biases.
Our key contributions can be summarized as follows:

\begin{itemize}
    \item We identify three novel types of scoring bias: \textbf{rubric order bias}, \textbf{score ID bias}, and \textbf{reference answer score bias}. To facilitate their evaluation, we introduce an automatic data synthesis pipeline that expands existing benchmarks into a tailored corpus, establishing a foundational resource for future research.
    
    \item We propose the first evaluation framework for investigating and quantifying scoring bias, introducing metrics across three key dimensions: \textbf{stability}, \textbf{accuracy}, and \textbf{scoring tendency}. 
    
    \item We empirically demonstrate that even the most advanced LLMs suffer from the substantial scoring biases we identify. Furthermore, our analysis yields valuable recommendations for designing more robust scoring prompts and mitigating these biases.
\end{itemize}

\section{Related Work}

\subsection{LLM-as-a-Judge}

Previous work~\cite{gu2024survey} defines LLM-as-a-Judge as the task in which an LLM generates an evaluation result in any form given input data and context.
Common forms of LLM judgment include score generation~\cite{gao2023human}, pairwise comparison~\cite{qin2024large}, ranking~\cite{li2023prd}, multiple-choice selection~\cite{li2024dalk}, and critique generation~\cite{ke2024critiquellm}.
Improvement strategies for LLM-as-a-Judge include meticulous design of evaluation prompts~\cite{liu2023g}, improvement of LLM's abilities such as fine-tuning~\cite{wangpandalm}, multi-LLM collaboration pipelines such as feedback mechanisms~\cite{xu2023instructscore}, and output optimization strategies such as multi-output ensembling~\cite{sottana2023evaluation}.
The performance of LLM-as-a-Judge is usually measured using metrics that represent its alignment with human annotators or advanced LLM judgments such as agreement~\cite{thakur2024judging}, Cohen's Kappa, and Spearman's correlation~\cite{bai2023benchmarking}.
LLM-as-a-Judge is widely applied to evaluation~\cite{lin2023llm}, alignment~\cite{bai2022constitutional}, retrieval~\cite{sun2023chatgpt}, reasoning~\cite{gao2024strategyllm}, and other specific domains~\cite{babaei2024gpt,ryu2023retrieval}.
LLM-as-a-Judge faces challenges of bias and vulnerability, which significantly compromise the fairness and reliability of LLM judges~\cite{li2024generation}.

\subsection{Bias in LLM-as-a-Judge}

Existing work~\cite{ye2024justice} defines the prompt for LLM judges as three components: system instruction, question, and response.
Based on this, they define bias in LLM-as-a-Judge as the tendency for judgments to differ when the system instruction or responses are modified in a bias-related manner.
Common types of bias in LLM-as-a-Judge include position bias, length bias, self-enhancement bias, etc.
These biases are often incorporated into the evaluation scope of existing LLM-as-a-Judge benchmarks~\cite{wei2024systematic,zheng2023judging}.
Currently, the types of bias remain a focal point of investigation, and numerous benchmarks have been proposed to evaluate bias in LLM-as-a-Judge.
OffsetBias~\cite{park2024offsetbias} identifies six types of bias in LLM-as-a-Judge including length bias, concreteness bias, empty reference bias, and so on.
CALM~\cite{ye2024justice} quantifies and analyzes 12 types of bias in LLM judges.
However, the biases defined in current work are predominantly focused on comparative evaluation tasks, and there is a lack of systematic research and definitions specifically addressing biases in scoring-based evaluation tasks.
Similar to research on adversarial attacks, the evaluation metrics for bias in LLM-as-a-Judge are often based on the consistency between bias-related modified judgments and the original judgments, or on alignment with golden answers.

Some studies focus on mitigating bias in LLM-as-a-Judge and aim to develop more robust LLM judges.
\cite{liualigning} introduces Pairwise-Preference Search (PAIRS) to mitigate biases in LLM evaluators.
\cite{wang2024large} proposes three methods to calibrate positional bias, including multiple evidence calibration, balanced position calibration, and human-in-the-loop calibration.
Our evaluation results on scoring bias will serve as a reference for future research on mitigating scoring biases in judge models and enhancing the robustness of scoring-based evaluations.

\section{Evaluation Method}

\begin{figure}[!t]
    \centering
    \includegraphics[width=0.85\linewidth]{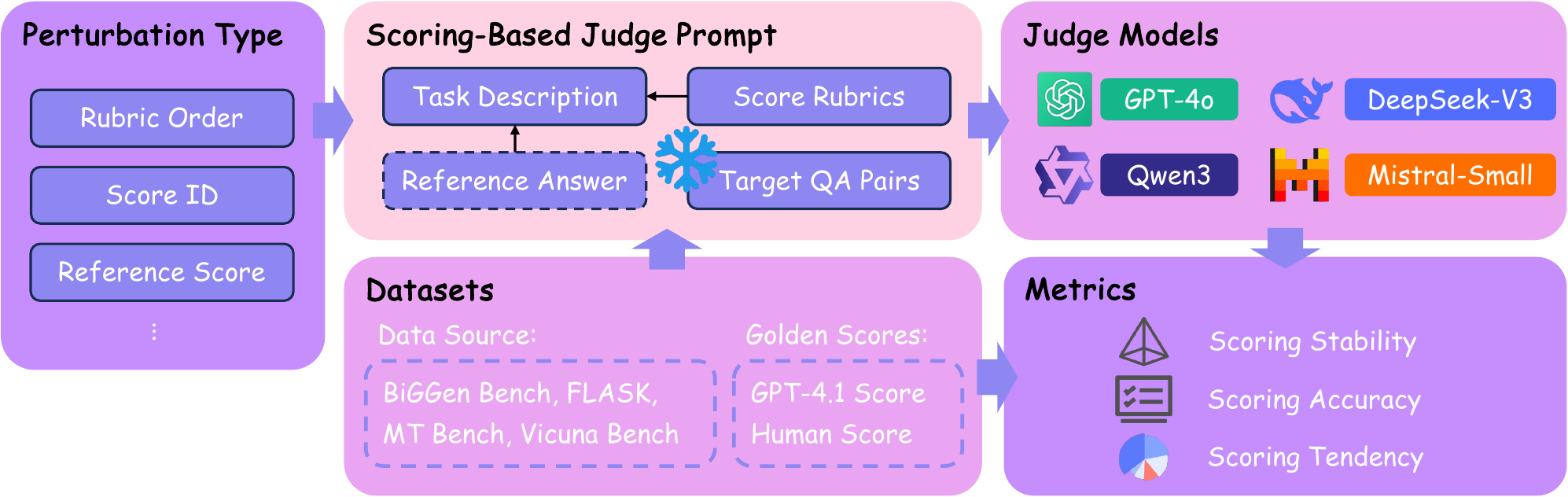}
    \caption{Our evaluation framework to measure the scoring bias in LLM-as-a-Judge.}
    \label{fig:framework}
\end{figure}

Figure~\ref{fig:framework} shows our evaluation framework, inspired by~\cite{ye2024justice}. On a selected dataset, we inject predefined perturbation types into the scoring prompts for the LLM judges. Then we obtain the scores from a selected judge model. We evaluate the scoring bias by calculating the consistency between scores before and after injecting perturbations, and the correlations between the judge model's scores and the golden scores (obtained from advanced LLMs and human annotators). Furthermore, our metrics include score distribution, which is used to measure the impact of different types of bias on the scoring tendencies of LLM judges.

\subsection{Problem Formulation}

\begin{figure}[!t]
  \centering
  \input{figures/prompt}
  \caption{Scoring prompt template adapted from~\cite{kim2024prometheus}, with \textcolor{red}{red} text indicating components that were perturbed in our experiments.}
  \label{fig:prompt}
\end{figure}

We assess the scoring bias in LLM-as-a-Judge by perturbing the scoring prompt and measuring the resulting impact on judgments.
Following~\cite{kim2024prometheus}, the input prompt $P$ for the scoring-based LLM judge consists of four components: a task description $T$, an evaluation target instance $I$, which can be either an instruction-response or a question-answer pair,  reference answer(s) with corresponding score(s) $A$, and score rubrics $R$ as shown in Figure~\ref{fig:prompt}.
The input prompt $P$ is formulated as follows:

\begin{equation}
    P = (T, I^*, [A], R),
\end{equation}
where $*$ denotes the component will not be perturbed, and $[\cdot]$ means that the enclosed component is optional within the input prompt. 
This formulation is central to our study. To specifically explore biases induced by non-target components (i.e., $T$, $R$, and $[A]$), we hold the evaluation target instance $I$ constant, as denoted by $I^*$.
We also treat the reference answer $[A]$ as an optional element to explore its impact on scoring robustness, as it is not a mandatory part of the scoring process.

By perturbing the reference answer $A \rightarrow A'$ or the score rubrics $R \rightarrow R'$, we get a perturbed input prompt $P'$.
At the same time, the task description $T$ will also be modified to $T'$ in conjunction with the perturbed components to ensure the prompt remains coherent.
The score from the perturbed prompt will be:
\begin{equation}
    s' = \textbf{LLM}(P').
\end{equation}

For each target instance $I$, we have a golden score $s$ from advanced LLMs or human annotators.
If $s$ and $s'$ differ, it demonstrates that the LLM judge exhibits scoring bias on the given sample.

\subsection{Perturbation Types}

We propose three methods to perturb key parts of the judge prompt to introduce scoring biases:
\begin{itemize}
    \item \textbf{Score Rubric Order}: The ordering of the description for each score within the rubric. We explore \textit{Ascending-Numeric}  (score 1 to score 5), \textit{Descending-Numeric} (score 5 to score 1), and \textit{Random-Numeric} orders (random score rubrics listed in prompt).
    \item \textbf{Score ID}: Typically, scores are represented using Arabic numerals $\{1, 2, 3, 4, 5\}$ as IDs. In this work, we explore the use of \textit{Letter-Grades} $\{\text{E}, \text{D}, \text{C}, \text{B}, \text{A}\}$ and \textit{Roman-Numerals} $\{\text{i}, \text{ii}, \text{iii}, \text{iv}, \text{v}\}$ as scoring IDs.
    \item \textbf{Reference Answer Score}: Based on the finding from \cite{kim2023prometheus,kim2024prometheus} that reference answers improve scoring accuracy for LLM judges, our work addresses an unexplored question: whether attaching a specific score $k$ to these references, denoted as \textit{Ref-}$k$, could degrade scoring robustness.
\end{itemize}

\subsection{Datasets}

Following previous work~\cite{chiang2025tract,kim2024prometheus}, we select four LLM-as-a-Judge datasets for scoring bias evaluation:

\begin{itemize}
    \item \textbf{BiGGen Bench}~\cite{kim2024biggen}: This is a principled generation benchmark with instance-specific evaluation criteria. We use the non-multilingual portion of its test set, which contains 2,780 responses for 695 instances, all evaluated by human annotators.
    \item \textbf{FLASK}~\cite{ye2024flask}: This is a fine-grained evaluation benchmark with 200 test prompts, 12 score rubrics, and 2,001 responses.  It also includes scores labeled by human annotators.
    \item \textbf{MT Bench}~\cite{zheng2023judging}: This is a multi-turn chat benchmark with 80 test prompts. \cite{kim2023prometheus} write 80 hand-crafted score rubrics and generate 320 responses from WizardLM-13B~\cite{xu2024wizardlm}, Vicuna-13B~\cite{chiang2023vicuna}, Llama-2-Chat-13B~\cite{touvron2023llama}, and GPT-3.5-Turbo-0613~\cite{openai2023gpt35turbo}.
    \item \textbf{Vicuna Bench}~\cite{chiang2023vicuna}: This is a single-turn chat benchmark with 80 test prompts.\cite{kim2023prometheus} also write 80 hand-crafted score rubrics and generate 320 responses from WizardLM-13B, Vicuna-13B, Llama-2-Chat-13B, and GPT-3.5-Turbo-0613.
\end{itemize}

\begin{table}[!t]
\caption{Statistics of our evaluation datasets.}
\label{tab:anly}
\vspace{2mm}
\centering
\scalebox{0.75}{
\begin{tabular}{@{}cccc@{}}
\toprule
\textbf{Dataset}    & \textbf{Data Size \ } & \textbf{Human Judgment \ } & \textbf{Reference Answer} \\ \midrule
\textbf{BiGGen Bench} & 2,780              & \Checkmark               & \Checkmark                \\
\textbf{FLASK}        & 2,001              & \Checkmark               & \Checkmark                \\
\textbf{MT Bench}     & 320                & \XSolidBrush             & \XSolidBrush              \\
\textbf{Vicuna Bench} & 320                & \XSolidBrush             & \XSolidBrush              \\ \bottomrule
\end{tabular}
}
\end{table}

\begin{table}[!t]
\caption{Comparison of the alignment between our LLM judgment pipeline and human golden scores, where $\rho$ and $r$ represent Spearman's and Pearson's correlation coefficients, respectively.}
\label{tab:human}
\centering
\vspace{2mm}
\scalebox{0.75}{
\begin{tabular}{@{}lcccc@{}}
\toprule
\multirow{2}{*}{\textbf{Model (Source of score)}} & \multicolumn{2}{c}{\textbf{BiGGen Bench}}             & \multicolumn{2}{c}{\textbf{FLASK}}                    \\ \cmidrule(l){2-3}\cmidrule(l){4-5} 
                                & $\rho$ &  $r$ & $\rho$ & $r$ \\ \midrule
\textbf{GPT-4.1 (our pipeline)}                & \textbf{0.6048}            & \textbf{0.6360}            & \textbf{0.6401}            & 0.6823                     \\
\textbf{GPT-4o (our pipeline)}                 & 0.6047                     & 0.6340                     & 0.6340                     & 0.6557                     \\
\textbf{GPT-4 (original benchmark)}                  & 0.5741                     & 0.6045                     & 0.6207                     & \textbf{0.6862}            \\ \bottomrule
\end{tabular}
}
\end{table}

The detailed statistics of our selected datasets are shown in Table~\ref{tab:anly}.
Due to the cost of manual annotation, the scores from advanced LLMs are usually used as golden scores~\cite{kim2024biggen,ye2024flask}.
The instances in the above four datasets are pre-annotated with scores from GPT-4~\cite{openai2023gpt4}, which is out of date at the current time. 
Therefore, we use more recent models such as GPT-4.1~\cite{openai2025gpt41} and GPT-4o~\cite{openai2024gpt4o} to obtain golden scores. 
Inspired by~\cite{kim2023prometheus,kim2024prometheus}, we repeat the scoring process three times using a judge prompt that includes a reference answer assigned a score of 5, and then take a majority vote to obtain the final score from the LLM evaluators.
Table~\ref{tab:human} shows that the GPT-4.1 scores exhibit the highest correlation with human scores.
Throughout the remainder of our experiments, the scores generated by GPT-4.1 are treated as the golden LLM scores.

\begin{figure}[!t]
    \centering
    \includegraphics[width=0.9\linewidth]{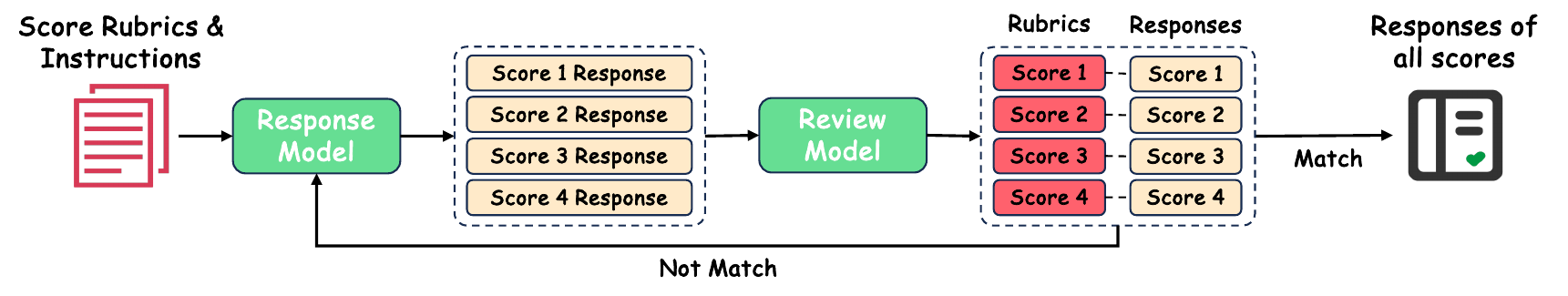}
    \caption{The proposed generation-review pipeline for synthesizing reference answers.}
    \label{fig:ref_ans_gen}
\end{figure}

Each instance in BiGGen Bench and FLASK is accompanied by a reference answer with a score of 5.
To measure the impact of the reference answer score on the judgment robustness, it is necessary to additionally generate reference answers with scores ranging from 1 to 4.
We propose a generation-review pipeline to synthesize reference answers for each sample in BiGGen Bench and FLASK, as shown in Figure~\ref{fig:ref_ans_gen}.
Given an instruction and score rubrics, the response model generates the response according to the description from score 1 to 4.
Then a review model checks whether the response matches the corresponding score description.
If they match, we accept the generated responses; otherwise, we rerun the pipeline until a match is achieved.
In this pipeline, we use GPT-4.1 and GPT-4o, alternating their roles as the response and review models.

\subsection{Metrics}

For a dataset of $n$ samples, the human-annotated golden label set is denoted as $S=\{s_1,\cdots,s_n\}$. We consider the ascending-numeric rubric without a reference answer as the \textbf{baseline} condition. The LLM outputs under this condition is denoted as $S^{(0)}=\{s_1^{(0)},\cdots,s_n^{(0)}\}$.
Let $\mathcal{P}$ be the set of all perturbed prompt templates and the unperturbed baseline, and $\mathcal{P}'$ be the set of all perturbed prompt templates excluding the baseline. For any prompt template $p$ in $\mathcal{P}$ or $\mathcal{P}'$, the resulting LLM score set is $S^{(p)}=\{s_1^{(p)},\cdots,s_n^{(p)}\}$.

\textbf{Stability metrics} quantify the consistency between the scores generated by a perturbed prompt $p, p \in \mathcal{P}'$ (i.e. $S^{(p)}$) and the baseline scores $S^{(0)}$. We adopt flip rate (FP) and mean absolute deviation (MAD) as metrics. FP and MAD are defined as follows,

\begin{align}
    \text{FP}^{(p)}  = \frac{1}{n} \sum_{i=1}^{n} \mathbb{I}(s_i^{(p)} \neq s_i^{(0)}), \quad
    \text{MAD}^{(p)}  = \frac{1}{n} \sum_{i=1}^{n} |s_i^{(p)} - s_i^{(0)}|, \quad
    p \in \mathcal{P}'.
\end{align}

\textbf{Accuracy metrics} quantify the agreement between the scores generated by a perturbed prompt $p, p \in \mathcal{P}$ (i.e. $S^{(p)}$) and the golden scores $S$. Specifically, we report both Spearman’s correlation coefficient $\rho$~\cite{spearman1904proof} and Pearson’s correlation coefficient $r$~\cite{pearson1895vii}.

\textbf{Scoring tendency} quantifies the model's tendency to favor specific score categories for a given prompt template $p$ ($p \in \mathcal{P}$). For rating scale $\{r_1,\cdots,r_5\}$ (e.g., $\{1,\cdots,5\}$), the tendency is defined by the following scoring distribution,
\begin{equation}
    N^{(p)} = (N_{r_1}^{(p)}, \cdots, N_{r_5}^{(p)}), \quad 
    N_{r_k}^{(p)} = \sum_{i=1}^{n}{\mathbb{I}(s_i^{(p)}=r_k)}, \quad 
    k \in \{1, \cdots, 5\}.
\end{equation}

\section{Experiments}

\subsection{Setup}

\textbf{Judge Models.}
Previous work~\cite{gao2024best,li2024quantifying,liu2024alignbench} has reached a consensus that judge models should be stronger models to ensure evaluation reliability~\cite{ye2024justice}.
After preliminary experiments, we found that scoring bias is a common phenomenon in judge models based on various LLMs. 
Furthermore, we aim to explore the impact of factors such as model type and size on scoring bias.
Therefore, we select diverse judge models that include closed-source, open-source, large-size, and small-size models.
They are GPT-4o, DeepSeek-V3-671B~\cite{liu2024deepseek}, Qwen3-32B, Qwen3-8B~\cite{yang2025qwen3}, and Mistral-Small-24B-Instruct-2501~\cite{mistralai2025mistral}.
These judge models we selected differ from the models used to generate responses in the evaluation datasets, thus mitigating potential self-enhancement bias~\cite{ye2024justice,zheng2023judging}.

\textbf{Judge Prompt.}
The prompt we use for scoring-based judgment, adapted from~\cite{kim2024prometheus}, is shown in Figure~\ref{fig:prompt}.
The unperturbed prompt is the one with ascending score rubrics, Arabic numeral score IDs, and no reference answer.

\textbf{Hyper-parameters.}
To ensure a deterministic scoring process, which is crucial for a fair evaluation of bias, we set the temperature of the LLM judge to 0 during all scoring tasks.

\subsection{Main Results}

\begin{table}[!t]
\caption{Stability metrics FP (\%) and MAD, comparing baseline scores with scores under various prompt perturbations. A \textbf{higher} metric indicates a greater difference from the baseline score. The perturbation method \textit{Ref-5} was only applied to BiGGen Bench and FLASK, as the other two datasets do not provide reference answers.}
\label{tab:flip_metrics}
\vspace{2mm}
\centering
\scalebox{0.7}{
\begin{tabular}{@{}cccccccccc@{}}
\toprule
\multirow{2}{2cm}{\textbf{Model}}         & \multirow{2}{*}{\textbf{Perturbation}} & \multicolumn{2}{c}{\textbf{BiGGen Bench}} & \multicolumn{2}{c}{\textbf{FLASK}} & \multicolumn{2}{c}{\textbf{MT Bench}} & \multicolumn{2}{c}{\textbf{Vicuna Bench}} \\ \cmidrule(l){3-4}\cmidrule(l){5-6}\cmidrule(l){7-8}\cmidrule(l){9-10} 
                                        &                                        & FR             & MAD                 & FR         & MAD              & FR           & MAD               & FR             & MAD                 \\ \midrule
\multirow{5}{2cm}{\textbf{GPT-4o}}        & \textit{Descending-Numeric}            & 23.63               & 0.2885              & 23.44           & 0.2769           & 23.13             & 0.2594            & \textbf{20.00}      & \textbf{0.2094}     \\
                                        & \textit{Random-Numeric}                & 22.55               & 0.2651              & 24.94           & 0.3023           & \textbf{28.44}    & \textbf{0.3375}   & 19.38               & 0.1938              \\
                                        & \textit{Letter-Grades}                 & 18.17               & 0.2122              & 18.24           & 0.2259           & 19.69             & 0.2188            & 19.38               & 0.2031              \\
                                        & \textit{Roman-Numerals}                & 16.51               & 0.1892              & 15.94           & 0.2014           & 23.13             & 0.2688            & 15.63               & 0.1625              \\
                                        & \textit{Ref-5}                         & \textbf{45.54}      & \textbf{0.5604}     & \textbf{45.58}  & \textbf{0.7166}  & -                 & -                 & -                   & -                   \\ \midrule
\multirow{5}{2cm}{\textbf{DeepSeek-V3-671B}}   & \textit{Descending-Numeric}            & 24.21               & 0.2867              & 21.89           & 0.2734           & 25.00             & 0.2969            & 27.50               & 0.2906              \\
                                        & \textit{Random-Numeric}                & 30.86               & 0.3565              & 32.83           & 0.4123           & \textbf{32.19}    & \textbf{0.3875}   & \textbf{39.06}      & \textbf{0.4250}     \\
                                        & \textit{Letter-Grades}                 & 19.86               & 0.2349              & 18.14           & 0.2374           & 17.81             & 0.1969            & 26.25               & 0.2844              \\
                                        & \textit{Roman-Numerals}                & 19.53               & 0.2338              & 17.89           & 0.2169           & 20.00             & 0.2344            & 25.00               & 0.2625              \\
                                        & \textit{Ref-5}                         & \textbf{38.99}      & \textbf{0.4863}     & \textbf{45.13}  & \textbf{0.6742}  & -                 & -                 & -                   & -                   \\ \midrule
\multirow{5}{2cm}{\textbf{Qwen3-32B}}     & \textit{Descending-Numeric}            & 28.56               & 0.3281              & 25.54           & 0.2989           & 29.38             & 0.3656            & \textbf{34.06}      & \textbf{0.3688}     \\
                                        & \textit{Random-Numeric}                & 26.12               & 0.2957              & 29.64           & 0.3448           & 33.44             & 0.3938            & 27.81               & 0.2906              \\
                                        & \textit{Letter-Grades}                 & 25.11               & 0.2770              & 24.34           & 0.2909           & 28.44             & 0.3313            & 29.06               & 0.3000              \\
                                        & \textit{Roman-Numerals}                & 28.92               & 0.3259              & 28.39           & 0.3273           & \textbf{51.88}    & \textbf{0.6875}   & 30.94               & 0.3156              \\
                                        & \textit{Ref-5}                         & \textbf{37.59}      & \textbf{0.7674}     & \textbf{44.58}  & \textbf{0.6512}  & -                 & -                 & -                   & -                   \\ \midrule
\multirow{5}{2cm}{\textbf{Qwen3-8B}}      & \textit{Descending-Numeric}            & \textbf{46.22}      & \textbf{0.5296}     & 32.23           & 0.3778           & \textbf{40.31}    & \textbf{0.4594}   & \textbf{44.06}      & \textbf{0.4750}     \\
                                        & \textit{Random-Numeric}                & 34.93               & 0.3989              & 36.83           & 0.4363           & 35.31             & 0.4125            & 34.38               & 0.3750              \\
                                        & \textit{Letter-Grades}                 & 30.54               & 0.3493              & 28.39           & 0.3228           & 36.88             & 0.4063            & 28.44               & 0.2938              \\
                                        & \textit{Roman-Numerals}                & 32.99               & 0.3827              & 34.38           & 0.4058           & \textbf{40.31}    & 0.4344            & 30.94               & 0.3156              \\
                                        & \textit{Ref-5}                         & 36.48               & 0.4367              & \textbf{40.53}  & \textbf{0.5712}  & -                 & -                 & -                   & -                   \\ \midrule
\multirow{5}{2cm}{\textbf{Mistral-Small-24B-Instruct}} & \textit{Descending-Numeric}            & 31.04               & 0.3421              & 27.39           & 0.2904           & \textbf{32.81}    & \textbf{0.3406}   & \textbf{25.63}      & \textbf{0.2719}     \\
                                        & \textit{Random-Numeric}                & 28.02               & 0.3083              & 33.03           & 0.3603           & 29.38             & 0.3066            & 22.81               & 0.2406              \\
                                        & \textit{Letter-Grades}                 & 25.36               & 0.2734              & 25.54           & 0.2809           & 26.88             & 0.2969            & 23.75               & 0.2438              \\
                                        & \textit{Roman-Numerals}                & 21.83               & 0.2349              & 23.24           & 0.2549           & 21.56             & 0.2281            & 16.56               & 0.1688              \\
                                        & \textit{Ref-5}                         & \textbf{35.40}      & \textbf{0.4065}     & \textbf{48.58}  & \textbf{0.6247}  & -                 & -                 & -                   & -                   \\ \bottomrule
\end{tabular}
}

\end{table}

\begin{table}[!t]
\caption{Accuracy metrics Spearman’s ($\rho$) and Pearson’s ($r$) correlation coefficient, showing the correlation between judge scores under perturbations and golden scores. Each judge model's first row serves as the unperturbed baseline. Results are colored \textcolor{green}{green} (better than baseline) or \textcolor{red}{red} (worse).}
\vspace{2mm}
\centering
\scalebox{0.7}{
\begin{tabular}{@{}cccccccccc@{}}
\toprule
\multirow{3}{2cm}{\textbf{Model}}         & \multirow{3}{*}{\textbf{Perturbation}}       & \multicolumn{4}{c}{\textbf{Human Score}}                                                                                                                                              & \multicolumn{4}{c}{\textbf{GPT-4.1 Score}}                                                                                                                                            \\ \cmidrule(l){3-6}\cmidrule(l){7-10}
                                        &                                              & \multicolumn{2}{c}{BiGGen Bench}                                                          & \multicolumn{2}{c}{FLASK}                                                                 & \multicolumn{2}{c}{MT Bench}                                                              & \multicolumn{2}{c}{Vicuna Bench}                                                          \\ \cmidrule(l){3-4}\cmidrule(l){5-6}\cmidrule(l){7-8}\cmidrule(l){9-10} 
                                        &                                              & $\rho$                                      & $r$                                         & $\rho$                                      & $r$                                         & $\rho$                                      & $r$                                         & $\rho$                                      & $r$                                         \\ \midrule
\multirow{5}{2cm}{\textbf{GPT-4o}}       & -                                            & 0.5669                                      & 0.6147                                      & 0.5529                                      & 0.5915                                      & 0.7578                                      & 0.7724                                      & 0.6862                                      & 0.7696                                      \\
                                        & \textit{Descending-Numeric} & \cellcolor{green!25}0.5836 & \cellcolor{green!25}0.6227 & \cellcolor{green!25}0.5754 & \cellcolor{green!25}0.6080 & \cellcolor{red!25}0.7479   & \cellcolor{red!25}0.7597   & \cellcolor{red!25}0.6734   & \cellcolor{red!25}0.7635   \\
                                        & \textit{Random-Numeric}     & \cellcolor{green!25}0.5947 & \cellcolor{green!25}0.6296 & \cellcolor{green!25}0.5592 & \cellcolor{red!25}0.5894   & \cellcolor{green!25}0.7673 & \cellcolor{green!25}0.7838 & \cellcolor{red!25}0.6701   & \cellcolor{red!25}0.7504   \\
                                        & \textit{Letter-Grades}      & \cellcolor{red!25}0.5627   & \cellcolor{red!25}0.6057   & \cellcolor{green!25}0.5558 & \cellcolor{green!25}0.5977 & \cellcolor{red!25}0.7317   & \cellcolor{red!25}0.7454   & \cellcolor{red!25}0.6539   & \cellcolor{green!25}0.7712 \\
                                        & \textit{Roman-Numerals}     & \cellcolor{green!25}0.5753 & \cellcolor{green!25}0.6172 & \cellcolor{green!25}0.5535 & \cellcolor{red!25}0.5912   & \cellcolor{green!25}0.7725 & \cellcolor{green!25}0.7786 & \cellcolor{green!25}0.7073 & \cellcolor{green!25}0.7821 \\ \midrule
\multirow{5}{2cm}{\textbf{DeepSeek-V3-671B}}   & -                                            & 0.5340                                      & 0.5823                                      & 0.5193                                      & 0.5746                                      & 0.6323                                      & 0.6593                                      & 0.4411                                      & 0.6119                                      \\
                                        & \textit{Descending-Numeric} & \cellcolor{green!25}0.5383 & \cellcolor{green!25}0.5890 & \cellcolor{green!25}0.5212 & \cellcolor{red!25}0.5642   & \cellcolor{red!25}0.6184   & \cellcolor{red!25}0.6436   & \cellcolor{green!25}0.5106 & \cellcolor{green!25}0.6656 \\
                                        & \textit{Random-Numeric}     & \cellcolor{red!25}0.5161   & \cellcolor{red!25}0.5685   & \cellcolor{red!25}0.5029   & \cellcolor{red!25}0.5560   & \cellcolor{red!25}0.6178   & \cellcolor{red!25}0.6460   & \cellcolor{green!25}0.5178 & \cellcolor{green!25}0.6187 \\
                                        & \textit{Letter-Grades}      & \cellcolor{green!25}0.5410 & \cellcolor{green!25}0.5835 & \cellcolor{green!25}0.5350 & \cellcolor{green!25}0.5840 & \cellcolor{green!25}0.6489 & \cellcolor{green!25}0.6715 & \cellcolor{green!25}0.4938 & \cellcolor{green!25}0.6322 \\
                                        & \textit{Roman-Numerals}     & \cellcolor{red!25}0.5184   & \cellcolor{red!25}0.5731   & \cellcolor{red!25}0.5090   & \cellcolor{red!25}0.5624   & \cellcolor{green!25}0.6358 & \cellcolor{green!25}0.6642 & \cellcolor{green!25}0.4504 & \cellcolor{red!25}0.5897   \\ \midrule
\multirow{5}{2cm}{\textbf{Qwen3-32B}}     & -                                            & 0.5418                                      & 0.5848                                      & 0.5323                                      & 0.5695                                      & 0.6959                                      & 0.7053                                      & 0.6419                                      & 0.7188                                      \\
                                        & \textit{Descending-Numeric} & \cellcolor{green!25}0.5572 & \cellcolor{green!25}0.5989 & \cellcolor{green!25}0.5345 & \cellcolor{red!25}0.5657   & \cellcolor{red!25}0.6843   & \cellcolor{red!25}0.6989   & \cellcolor{red!25}0.6205   & \cellcolor{red!25}0.7020   \\
                                        & \textit{Random-Numeric}     & \cellcolor{green!25}0.5528 & \cellcolor{green!25}0.5938 & \cellcolor{red!25}0.5115   & \cellcolor{red!25}0.5456   & \cellcolor{red!25}0.6554   & \cellcolor{red!25}0.6688   & \cellcolor{red!25}0.6416   & \cellcolor{green!25}0.7252 \\
                                        & \textit{Letter-Grades}      & \cellcolor{red!25}0.5352   & \cellcolor{green!25}0.5852 & \cellcolor{green!25}0.5393 & \cellcolor{green!25}0.5724 & \cellcolor{red!25}0.6795   & \cellcolor{red!25}0.6930   & \cellcolor{red!25}0.6208   & \cellcolor{red!25}0.7079   \\
                                        & \textit{Roman-Numerals}     & \cellcolor{red!25}0.5139   & \cellcolor{red!25}0.5629   & \cellcolor{red!25}0.5257   & \cellcolor{green!25}0.5781 & \cellcolor{green!25}0.7158 & \cellcolor{green!25}0.7238 & \cellcolor{red!25}0.6072   & \cellcolor{red!25}0.7126   \\ \midrule
\multirow{5}{2cm}{\textbf{Qwen3-8B}}      & -                                            & 0.5087                                      & 0.5419                                      & 0.5035                                      & 0.5309                                      & 0.6238                                      & 0.6342                                      & 0.6639                                      & 0.7284                                      \\
                                        & \textit{Descending-Numeric} & \cellcolor{green!25}0.5223 & \cellcolor{green!25}0.5592 & \cellcolor{red!25}0.4885   & \cellcolor{green!25}0.5339 & \cellcolor{red!25}0.6005   & \cellcolor{red!25}0.6135   & \cellcolor{red!25}0.5500   & \cellcolor{red!25}0.6647   \\
                                        & \textit{Random-Numeric}     & \cellcolor{red!25}0.4831   & \cellcolor{red!25}0.5107   & \cellcolor{red!25}0.4778   & \cellcolor{red!25}0.5004   & \cellcolor{red!25}0.5949   & \cellcolor{red!25}0.5983   & \cellcolor{red!25}0.5135   & \cellcolor{red!25}0.6053   \\
                                        & \textit{Letter-Grades}      & \cellcolor{red!25}0.4842   & \cellcolor{red!25}0.5195   & \cellcolor{red!25}0.4873   & \cellcolor{red!25}0.5244   & \cellcolor{red!25}0.5985   & \cellcolor{red!25}0.6126   & \cellcolor{red!25}0.5812   & \cellcolor{red!25}0.6867   \\
                                        & \textit{Roman-Numerals}     & \cellcolor{red!25}0.4808   & \cellcolor{red!25}0.5165   & \cellcolor{red!25}0.4606   & \cellcolor{red!25}0.5087   & \cellcolor{red!25}0.6016   & \cellcolor{red!25}0.5975   & \cellcolor{red!25}0.6111   & \cellcolor{red!25}0.7178   \\ \midrule
\multirow{5}{2cm}{\textbf{Mistral-Small-24B-Instruct}} & -                                            & 0.5067                                      & 0.5543                                      & 0.4948                                      & 0.5392                                      & 0.5943                                      & 0.5762                                      & 0.6116                                      & 0.6917                                      \\
                                        & \textit{Descending-Numeric} & \cellcolor{green!25}0.5383 & \cellcolor{green!25}0.5767 & \cellcolor{red!25}0.4921   & \cellcolor{red!25}0.5325   & \cellcolor{green!25}0.6104 & \cellcolor{green!25}0.6102 & \cellcolor{green!25}0.6137 & \cellcolor{green!25}0.7082 \\
                                        & \textit{Random-Numeric}     & \cellcolor{red!25}0.5047   & \cellcolor{red!25}0.5450   & \cellcolor{red!25}0.4758   & \cellcolor{red!25}0.5065   & \cellcolor{red!25}0.5603   & \cellcolor{red!25}0.5602   & \cellcolor{red!25}0.5461   & \cellcolor{red!25}0.6418   \\
                                        & \textit{Letter-Grades}      & \cellcolor{green!25}0.5098 & \cellcolor{green!25}0.5610 & \cellcolor{red!25}0.4894   & 0.5392                                      & \cellcolor{red!25}0.5375   & \cellcolor{red!25}0.5485   & \cellcolor{red!25}0.5558   & \cellcolor{red!25}0.6608   \\
                                        & \textit{Roman-Numerals}     & \cellcolor{green!25}0.5107 & \cellcolor{green!25}0.5612 & \cellcolor{red!25}0.4932   & \cellcolor{green!25}0.5455 & \cellcolor{red!25}0.5863   & \cellcolor{green!25}0.5907 & \cellcolor{red!25}0.5903   & \cellcolor{red!25}0.6905   \\ \bottomrule
\end{tabular}}

\label{tab:corr_golden_main}
\end{table}

\textbf{The existence of scoring bias is proved in all judge models.}
Table~\ref{tab:flip_metrics} shows the difference between baseline model scores and perturbed model scores, Table~\ref{tab:corr_golden_main} and Figure~\ref{fig:ref_ans_cor} show the correlation between judge model scores and golden scores under all perturbation methods.
For all judge models we evaluated, including advanced models such as GPT-4o, we observed fluctuations in the scoring results under different types of perturbations. This indicates that scoring bias significantly impacts the robustness of LLM-as-a-Judge.
Furthermore, different judge models exhibit varying sensitivity to scoring bias:

\textbf{(1) More powerful judge models tend to be more robust to scoring bias.}
Consistent with existing consensus~\cite{gao2024best,li2024quantifying,liu2024alignbench}, our experimental results indicate that judge models with advanced capability and larger size tend to exhibit higher reliability in scoring.
As shown in Table~\ref{tab:flip_metrics}, GPT-4o has FRs less than 25\% and MADs less than 0.3 under perturbation in most cases, except for the case involving a full-marked reference answer.
Althogh Qwen3-8B, has higher FRs and MADs than GPT-4o and the larger version Qwen3-32B, especially reaches 46.22\% FR and 0.5296 MAD on BiGGen Bench under descending-numeric perturbation, which means that nearly half of the scores are corrupted.
Therefore, more capable judge models should be prioritized in scoring evaluations.

\textbf{(2) Contrary to the conventional view that bias is always harmful, we find that scoring bias in LLM-as-a-Judge can also have surprisingly positive effects on the evaluation results.}
For example, as Table~\ref{tab:corr_golden_main} shows, in most cases, the correlation between GPT-4o scores and golden scores increases when Roman numerals are used as scoring IDs.
However, when DeepSeek-V3-671B, Qwen3-32B and Qwen3-8B are employed as judge models, the use of Roman numerals generally exerts a negative effect on their scoring accuracy.
\textit{An interesting phenomenon is that the impact of reference answer score bias on the judge models we evaluated appears to be consistent, which will be further analyzed in subsequent sections}.
Our experimental results can serve as a valuable reference for the selection of judge models and the design of scoring prompt templates.

\begin{figure}[!t]
    \centering
    \begin{subfigure}[b]{0.51\textwidth}
        \includegraphics[width=\textwidth]{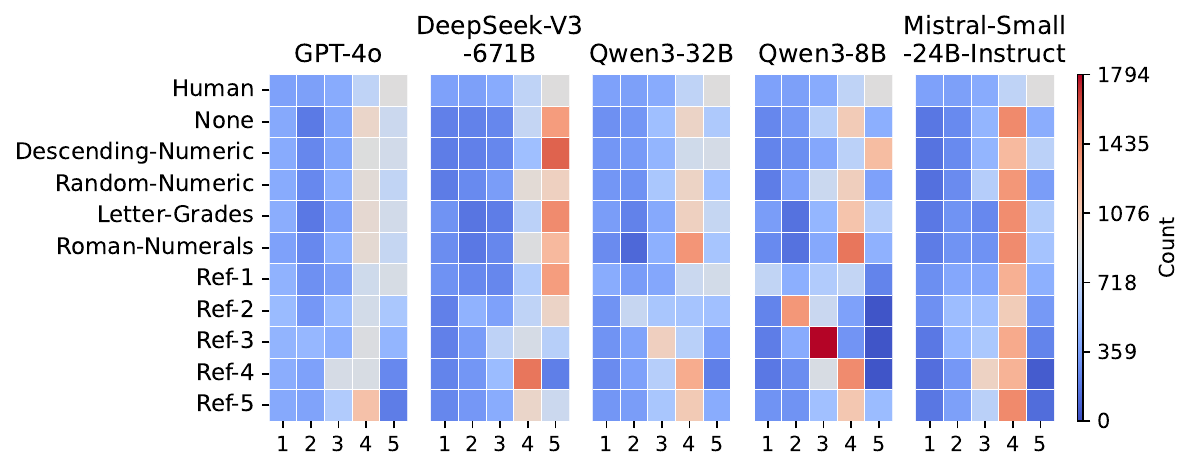}
        \caption{BiGGen Bench}
        \label{fig:heat_map_1}
    \end{subfigure}
    \hspace{-5mm}
    \begin{subfigure}[b]{0.51\textwidth}
        \includegraphics[width=\textwidth]{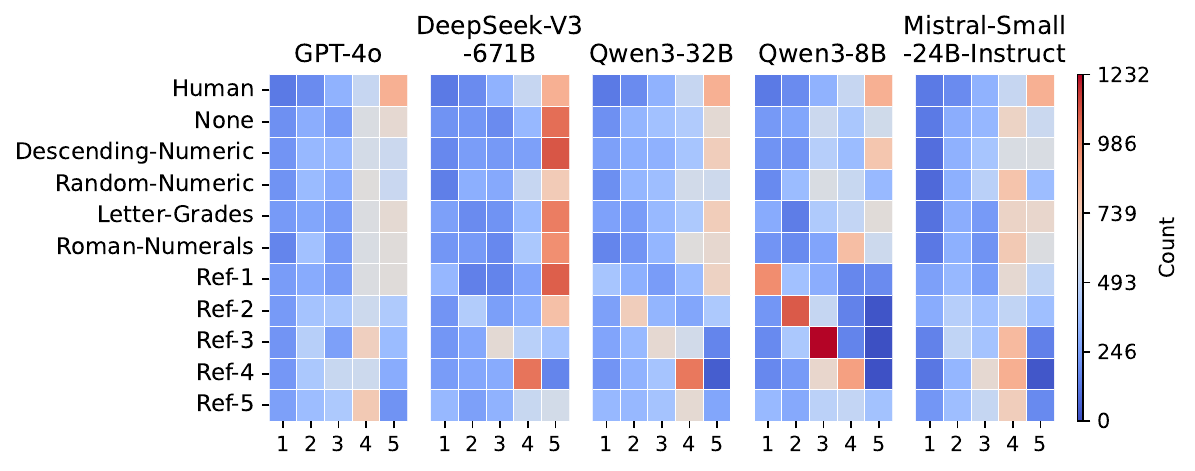}
        \caption{FLASK}
        \label{fig:heat_map_2}
    \end{subfigure}
    \caption{Score tendency of LLM judges on BiGGen Bench and FLASK, plotting score distributions (x-axis) against perturbation methods (y-axis). The ``Human'' (top) and ``None'' (second) rows serve as baselines, showing human scores and the model's unperturbed scores, respectively. }
    \label{fig:heat_map}
\end{figure}

\textbf{(3) Judge models have their own scoring tendencies.}
Figure~\ref{fig:heat_map} presents the scoring tendencies of different judge models under various scoring biases in the form of heatmaps.
It is obvious that each judge model exhibits its own specific scoring preferences.
Under unbiased conditions, DeepSeek-V3-671B assigns a score of 5 to more than half of the instruction-response pairs.
Compared to the human scoring distribution, Mistral-Small-24B-Instruct has a stronger tendency to assign a score of 4 to the samples.
Other judge models show smoother scoring distribution shifts, but certain scores are still preferred.
For example, GPT-4o demonstrates a preference for a score of 4.
An examination of Figure~\ref{fig:heat_map} reveals that score rubric order bias and score ID bias have a certain impact on the scoring tendencies of the judge models; however, they do not fundamentally alter the overall score distribution.
An exception is Qwen3-8B, whose score distribution is more significantly influenced by biases, possibly due to its small size.
It is particularly evident that the reference answer score bias exerts a substantial impact on the score distribution, which will be analyzed in detail in the following sections.
Our experimental results show that judge models inherently exhibit biases in their scoring tendencies.
Therefore, debiasing the judge models is crucial to ensuring the fairness and reliability of the evaluation.

\subsection{Methods for Scoring Bias Mitigation}

\subsubsection{(1) Leveraging Model Scale as a Debiasing Strategy.}

Our findings show a strong correlation between model scale and bias resilience. As shown in Table~\ref{tab:flip_metrics}, state-of-the-art models like GPT-4o consistently demonstrate high stability, with FRs often remaining below 25\%. In stark contrast, smaller models are highly vulnerable. Qwen3-8B, for example, suffers a 46.22\% FR on BiGGen Bench under a simple rubric reordering. The larger Qwen3-32B is far more stable (28.56\% FR) under the same perturbation, confirming the link to model scale.
This trend extends to accuracy, as shown in Table~\ref{tab:corr_golden_main}. Perturbations often improve GPT-4o's correlations with human scores, whereas the same perturbations frequently degrade the accuracy of smaller models like Qwen3-8B and Mistral-Small. Therefore, for high-stakes evaluations, \textbf{selecting a powerful model like GPT-4o is a primary strategy for ensuring robust and reliable results}, justifying potential trade-offs in cost or latency.

\subsubsection{(2) Debiasing via Intentional Deviation from Convention.}

In most cases of scoring-based evaluation tasks, Arabic numerals are commonly used as scoring IDs, and the score rubric order typically follows a sequential order (1 to 5). These conventional choices align with human cognitive preferences in prompt design. However, our experimental results (Table~\ref{tab:corr_golden_main}) indicate that such prompts do not necessarily produce the most accurate scoring. For example, the GPT-4o judge improves when using Roman numerals. Besides, using letter grades as scoring IDs improves the scoring accuracy of DeepSeek-V3-671B. Regarding the score rubric order bias, adopting a descending order improves the scoring accuracy of DeepSeek-V3 and Mistral-Small-24B-Instruct in most instances. Intuitively, employing a random scoring rubric order tends to negatively impact the scoring accuracy of all judge models. Based on our analysis, \textbf{we propose a core design principle for scoring prompts: intentional yet meaningful deviation from convention}. Effective strategies include structured, unconventional formats, such as presenting rubrics in descending order or using non-numeric identifiers like letter grades (A, B, C). However, the key is to maintain clarity. Therefore, chaotic perturbations—such as a fully randomized rubric order—should be avoided as they harm logical coherence.
Moreover, inspired by previous work~\cite{liualigning,park2024offsetbias}, perturbations applied to scoring prompt also show potential as debiasing strategies.

\subsubsection{(3) Using full-mark reference answers can stabilize scoring preferences and achieve optimal accuracy.}
\begin{figure}[!t]
    \centering
    \begin{subfigure}[b]{0.5\textwidth}
        \includegraphics[width=\textwidth]{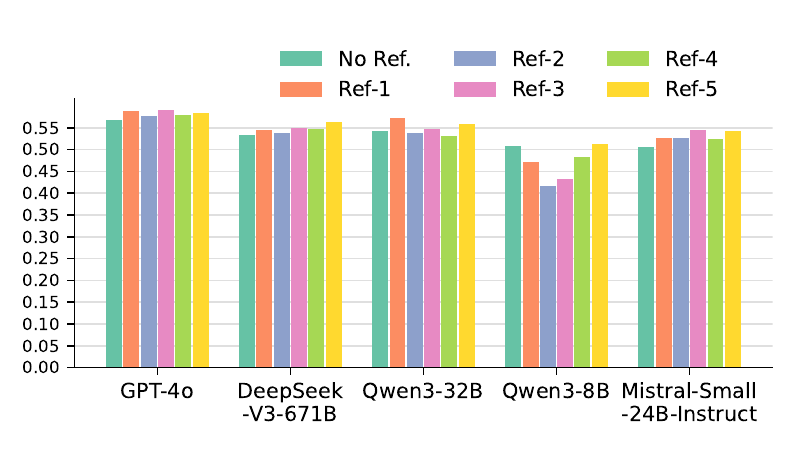}
        \caption{BiGGen Bench}
        \label{fig:ref_ans_cor_1}
    \end{subfigure}
    \hspace{-4mm}
    \begin{subfigure}[b]{0.50\textwidth}
        \includegraphics[width=\textwidth]{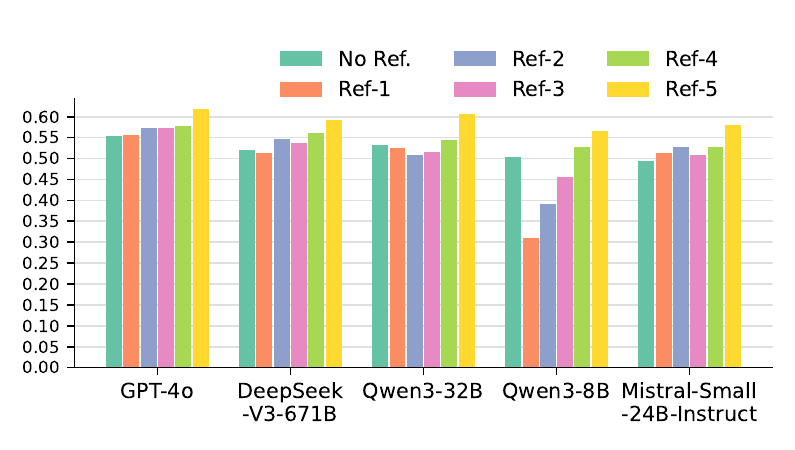}
        \caption{FLASK}
        \label{fig:ref_ans_cor_2}
    \end{subfigure}
    \caption{Accuracy metric Spearman's correlation between human scores and judge model scores on BiGGen Bench and FLASK dataset, under perturbations applied to the reference answer scores.}
    \label{fig:ref_ans_cor}
\end{figure}

The impact of reference answer score bias on scoring accuracy and tendency is shown in Figure~\ref{fig:ref_ans_cor} and Figure~\ref{fig:heat_map}. Our results indicate this bias's influence on robustness is more pronounced and regular than that of score rubric order or ID bias.

As shown in Figure~\ref{fig:ref_ans_cor}, introducing reference answers generally improves scoring accuracy for models like GPT-4o, DeepSeek-V3-671B, and Mistral-Small-24B-Instruct. Notably, across both BiGGen Bench and FLASK, introducing a full-marked (score 5) reference answer consistently yields top-tier scoring accuracy, significantly outperforming other scores and the baseline.

Figure~\ref{fig:heat_map} reveals a significant impact on scoring tendency, especially for DeepSeek-V3-671B, Qwen3-32B, and Qwen3-8B. When the reference score is between 1 and 4, these models tend to assign similar scores, indicating a strong pull from the reference. Interestingly, when a score 5 reference answer is introduced, the judge's scoring behavior becomes more rational.

In particular, when reference answers with a score of 5 are introduced, the models achieve the highest scoring accuracy without disproportionately influencing their scoring preferences. Therefore, \textbf{when incorporating reference answers into a scoring-based judge prompt, a full-mark reference answer is the best choice}. Conceptually, this aligns with metrics like BLEU and ROUGE, which also rely on golden answers.

\section{Limitations and Future Work}

While our work offers notable conclusions and insights, several limitations remain to be addressed in future work.

Firstly, our evaluation framework has room for further expansion.
More types of scoring biases remain to be identified.
Building more comprehensive benchmarks to evaluate scoring bias in LLM-as-a-Judge is also warranted, as a limited data distribution may lead to an unfair evaluation.

Secondly, mitigating scoring bias needs further exploration.
Although there are no dedicated methods for addressing scoring bias, intuitive and straightforward approaches, such as scoring multiple times followed by majority voting or averaging the scores, require empirical validation for their effectiveness in mitigating scoring bias.
Moreover, the elaborate design of scoring prompt templates, the construction of debiased scoring-based judgment datasets~\cite{park2024offsetbias}, and adversarial training are potential approaches to enhancing the scoring robustness of judge models.

Additionally, although our work demonstrates the existence of scoring bias in LLM-as-a-Judge, the underlying causes of scoring bias remain to be validated.
Approaches such as training data analysis and information flow observation~\cite{yuaninstance} may help identify the reasons for scoring bias, whether it originates from within the model or from external factors.

\section{Conclusion}

This work investigates \textbf{scoring bias} in LLM-as-a-Judge, where scoring judgments shift due to minor perturbations in judge prompts. To evaluate this, we designed a framework that perturbs score rubric order, score IDs, and reference answers, quantifying the bias using \textbf{scoring stability}, \textbf{scoring accuracy} and \textbf{scoring tendency}. Experiments on synthesized datasets expanded from four benchmarks confirm that scoring bias significantly undermines the robustness and reliability of existing LLM judges, with varying sensitivity across different models. To mitigate this, we recommend unconventional prompt designs—such as descending rubric order or using letter or Roman numeral IDs—and providing a full-mark reference answer when applicable, both of which effectively improve performance. In summary, our work provides evidence and actionable recommendations for evaluating and enhancing the reliability of LLM judges.

\begin{credits}

\subsubsection{\discintname}
The authors have no competing interests to declare that are relevant to the content of this article. 
\end{credits}

\bibliographystyle{splncs04}
\bibliography{mybibliography}

\end{document}